\documentclass[letterpaper]{article} % DO NOT CHANGE THIS
\usepackage{aaai25}  % DO NOT CHANGE THIS
\usepackage{times}  % DO NOT CHANGE THIS
\usepackage{helvet}  % DO NOT CHANGE THIS
\usepackage{courier}  % DO NOT CHANGE THIS
\usepackage[hyphens]{url}  % DO NOT CHANGE THIS
\usepackage{graphicx} % DO NOT CHANGE THIS
\usepackage{natbib}  % DO NOT CHANGE THIS AND DO NOT ADD ANY OPTIONS TO IT
\usepackage{caption} % DO NOT CHANGE THIS AND DO NOT ADD ANY OPTIONS TO IT
\usepackage{algorithm}
\usepackage{algorithmic}
\usepackage{graphicx}
\usepackage{booktabs}
\usepackage[algo2e]{algorithm2e} 
\usepackage{amsmath}
\usepackage{multirow} 
\usepackage{booktabs}       % professional-quality tables
\usepackage{amsfonts}       % blackboard math symbols
\usepackage{nicefrac}       % compact symbols for 1/2, etc.
\usepackage{microtype}      % microtypography
\usepackage{xcolor}         % colors

\usepackage{amsthm,amsmath,amssymb}
\usepackage{mathrsfs}
\usepackage{threeparttable}
\usepackage{bigstrut}
\usepackage{float}

\usepackage{newfloat}
\usepackage{listings}
\DeclareCaptionStyle{ruled}{labelfont=normalfont,labelsep=colon,strut=off} % DO NOT CHANGE THIS
\floatstyle{ruled}
\newfloat{listing}{tb}{lst}{}
\floatname{listing}{Listing}
\title{Zero-Shot Video Restoration and Enhancement Using Pre-Trained \\Image Diffusion Model}
\author{
    Cong Cao\textsuperscript{\rm 1},
    Huanjing Yue\textsuperscript{\rm 1}\thanks{Corresponding author},
    Xin Liu\textsuperscript{\rm 2},
    Jingyu Yang\textsuperscript{\rm 1}
}
\affiliations{
    \textsuperscript{\rm 1}School of Electrical and Information Engineering, Tianjin University, Tianjin, China\\
    \textsuperscript{\rm 2}Computer Vision and Pattern Recognition Laboratory, School of Engineering Science, Lappeenranta-Lahti University of Technology LUT, Lappeenranta, Finland\\
    caocong\_123@tju.edu.cn, huanjing.yue@tju.edu.cn, linuxsino@gmail.com, yjy@tju.edu.cn
}

\usepackage{bibentry}
\begin{document}

\maketitle

\begin{abstract}
% without training
Diffusion-based zero-shot image restoration and enhancement models have achieved great success in various tasks of image restoration and enhancement. However, directly applying them to video restoration and enhancement results in severe temporal flickering artifacts. In this paper, we propose the first framework for zero-shot video restoration and enhancement based on the pre-trained image diffusion model. By replacing the spatial self-attention layer with the proposed short-long-range (SLR) temporal attention layer, the pre-trained image diffusion model can take advantage of the temporal correlation between frames. We further propose temporal consistency guidance, spatial-temporal noise sharing, and an early stopping sampling strategy to improve temporally consistent sampling. Our method is a plug-and-play module that can be inserted into any diffusion-based image restoration or enhancement methods to further improve their performance. Experimental results demonstrate the superiority of our proposed method. 
\end{abstract}

\begin{links}
     \link{Code}{https://github.com/cao-cong/ZVRD}
\end{links}

\section{Introduction}
\label{sec:intro}

Recently, Denoising Diffusion Probabilistic Models (DDPMs) \cite{dhariwal2021diffusion} have demonstrated advanced generative capabilities surpassing those of GANs, inspiring further exploration of restoration and enhancement methods based on diffusion models. Different from using supervised learning and diffusion framework to train models for specific restoration and enhancement tasks \cite{saharia2022image,yin2023cle}, the works in \cite{song2019generative,chung2022diffusion,fei2023generative,shi2024conditional} employ a pre-trained image diffusion model for universal zero-shot image restoration and enhancement. These methods constrain the content between generated results in the reverse diffusion process and degraded images. However, due to the absence of temporal modeling in pre-trained image diffusion models, although these methods have shown promising results in image restoration and enhancement, their direct application to video restoration and enhancement can lead to significant temporal flickering.

With the emergence of powerful pre-trained text-to-image diffusion models, such as Stable Diffusion \cite{rombach2022high}, using off-the-shelf text-to-image diffusion model for zero-shot video editing has garnered increasing attention \cite{wu2023tune,yang2023rerender}. To generate temporally consistent edited video, the motion information from the original video is typically utilized to design various temporal modules \cite{cong2023flatten}. However, predicting motion becomes more challenging when dealing with video restoration and enhancement tasks since input videos suffering from various degradations. In order to address this issue, we propose Short-Long-Range (SLR) temporal attention which consists cross-neighbour-frame attention and self-corrected trajectory attention. The cross-neighbor-frame attention implicitly models short-range temporal correlation without explicitly estimating motion, while the self-corrected trajectory attention compensates for inaccurate explicit motion estimation to capture long-range temporal correlation. The explicitly estimated motion information is utilized to construct guidance for pixel-level temporal consistency, which is a complementary of semantic-level consistency guidance.
%works in conjunction with complementary semantic-level consistency guidance.
We observe that temporal flickering is mainly caused by inherent stochasticity in the diffusion model. Therefore, we introduce spatial-temporal noise sharing to mitigate this stochasticity effect. Additionally, we propose an early stopping sampling strategy since flicking details are usually generated during sampling in the later stage.

%according to the order of constructing low-frequency and high-frequency in a frame during sampling.
%In this paper, we propose a novel framework for Zero-shot Video Restoration and enhancement using a pretrained image Diffusion model (ZVRD).

%Our contributions are summarized as follows
%\begin{itemize}
%\item[$\bullet$]{First, we propose the first framework for zero-shot video restoration and enhancement using a pre-trained image diffusion model.}
%\item[$\bullet$]{Secondly, we propose SLR temporal attention, temporal consistency guidance, spatial-temporal noise sharing, and early stopping sampling strategy to maintain temporal consistency during video restoration and enhancement.}
%\item[$\bullet$]{Extensive experiments demonstrate the effectiveness of our method in achieving temporally consistent zero-shot video restoration and enhancement.}
%\end{itemize}
In summary, there are mainly three contributions in this work. First, we propose the first framework for Zero-shot Video Restoration and enhancement using a pre-trained image Diffusion model (ZVRD). Second, we propose SLR temporal attention, temporal consistency guidance, spatial-temporal noise sharing, and early stopping sampling strategy to maintain temporal consistency during video restoration and enhancement. Third, extensive experiments demonstrate the effectiveness of our method. 

%in achieving temporally consistent zero-shot video restoration and enhancement.

\section{Related Works}

\subsection{Diffusion-Based Zero-Shot Image Restoration and Enhancement}

The success of diffusion-based generative models has enlightened diffusion-based image restoration and enhancement methods. These methods can be divided into two categories. One category is designed for each specific task and utilizes paired data for supervised training \cite{saharia2022image,yin2023cle}. The other category is a universal zero-shot method for different image restoration tasks based on a pre-trained image diffusion model \cite{song2019generative,wang2022zero,chung2022diffusion,fei2023generative,shi2024conditional}. Zero-shot methods utilize a pre-trained off-the-shelf diffusion model as the generative prior, which requires no additional training. The key to zero-shot methods is to constrain the result in the reverse diffusion process to have consistent content as degraded images. DDNM \cite{wang2022zero} refines only the null-space contents during the reverse diffusion process to preserve content consistency. DPS \cite{chung2022diffusion} extends diffusion solvers to efficiently handle general noisy non-linear inverse problems via approximation of the posterior sampling. GDP \cite{fei2023generative} applies different loss functions between result and degraded image, and guides the reverse diffusion process with gradient. But these methods are designed for image restoration problems, there exists severe temporal flickering when applied to degraded videos.

\subsection{Diffusion-Based Zero-Shot Video Editing}

Along with the development of powerful pre-trained text-to-image diffusion models, such as Stable Diffusion \cite{rombach2022high}, diffusion-based zero-shot video editing has gained increasing attention, which utilizes the off-the-shelf text-to-image diffusion model and mainly solves the temporal consistency problem. FateZero \cite{qi2023fatezero} follows Prompt-to-Prompt \cite{hertz2022prompt} and fuses the attention maps to preserve the motion and structure consistency. Text2Video-Zero \cite{khachatryan2023text2video} proposes cross-frame attention for better temporal consistency. \cite{cong2023flatten,yang2024fresco} propose optical flow-guided attention and spatial-temporal correspondence-guided attention, respectively. Inspired by these works, we propose to use the pre-trained image diffusion model for zero-shot video restoration and enhancement. Different from these zero-shot video editing methods that use Stable Diffusion, we use an unconditional image diffusion model \cite{dhariwal2021diffusion} pre-trained on ImageNet, which is commonly used in zero-shot image restoration.

\subsection{Video Restoration and Enhancement}

The existing video restoration methods need to be trained for every single task.  Temporal mutual self-attention is proposed to exploit temporal information in video super-resolution and video deblurring \cite{liang2024vrt}. The work in \cite{yang2024bistnet} explores colors of exemplars and utilizes them to help video colorization by temporal feature fusion with the guidance of semantic image prior. The work in \cite{zheng2022semantic} explores zero-shot image (video) enhancement by utilizing non-reference loss functions, but still needs training on unpaired data with diverse illumination conditions. Different from the above methods, our method is a training-free zero-shot method, which is universal to different restoration and enhancement tasks. Recently, the work in \cite{yeh2024diffir2vr} adapts image restoration model for video restoration without training. But it is based on image latent diffusion model which has been specifically trained for restoration in a supervised manner. However, most zero-shot image restoration diffusion methods are based on \cite{dhariwal2021diffusion}. For the U-Net of \cite{dhariwal2021diffusion}, the attention module only exists on the features with $32\times32$ and lower resolution, the higher resolution features can also cause the temporal inconsistency of output. Therefore, besides the SLR temporal attention, we further propose temporal consistency guidance and spatial-temporal noise sharing to solve this problem. The token merging of the attention module in \cite{yeh2024diffir2vr} is not enough to maintain the temporal consistency of \cite{dhariwal2021diffusion}. Our method can be applied to both zero-shot and supervised diffusion-based image restoration (enhancement) models for video restoration (enhancement).

\begin{figure*}
    \centering
    \includegraphics[width=0.8\linewidth]{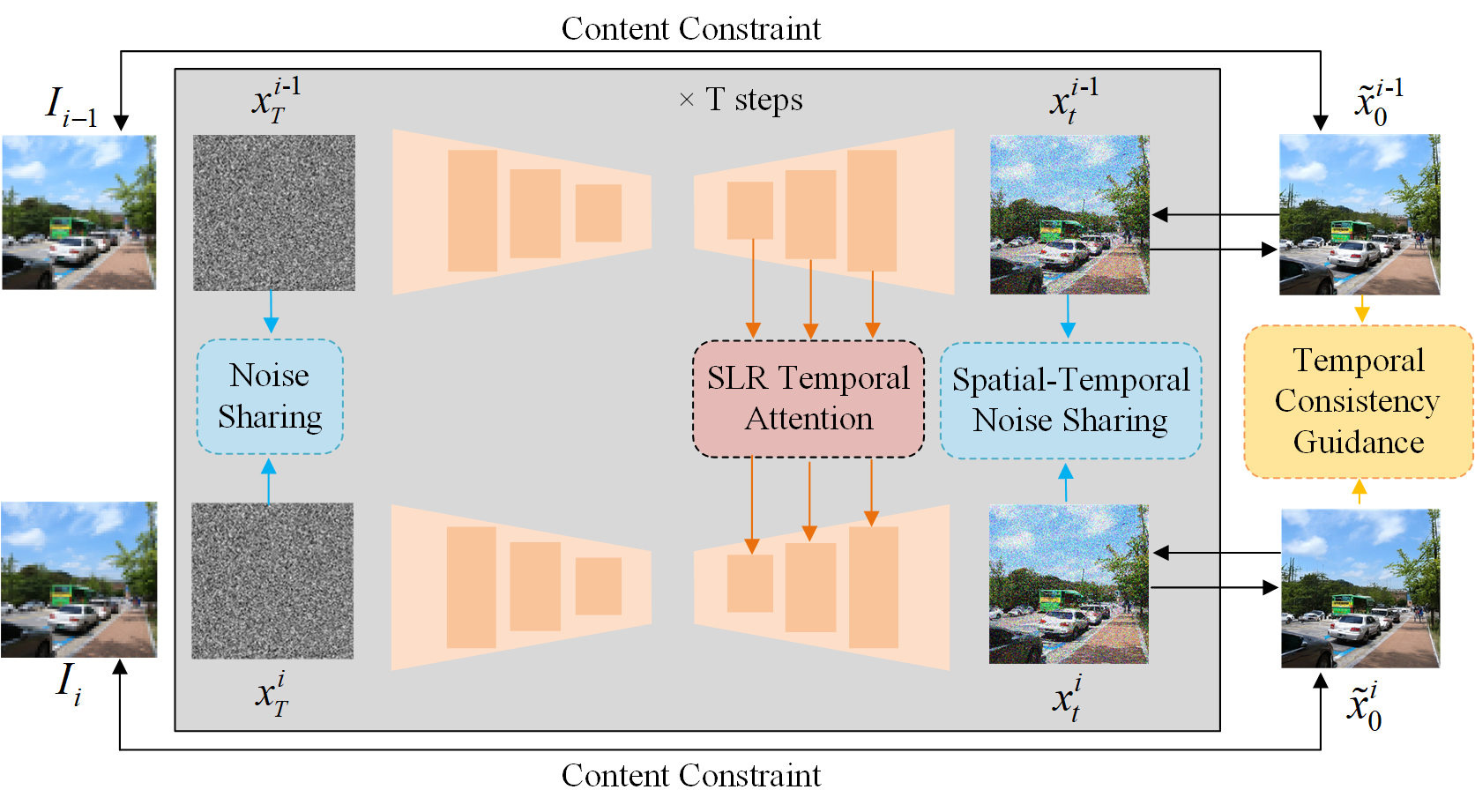}
    \caption{Framework of the proposed zero-shot video restoration and enhancement.}
    \label{fig:framework}
\end{figure*}

\section{Background}

Diffusion models transform target data distribution into simple noise distribution and recover data from noise. We follow the diffusion model defined in denoising diffusion probabilistic models (DDPM) \cite{ho2020denoising}. DDPM defines a T-step forward process and a T-step reverse process. The forward process adds random noise to data step by step, while the reverse process constructs target data samples step by step.

% The diffusion process from clean data $\boldsymbol{x}_0$ to $\boldsymbol{x}_T$ is written as:

% \lyu{always use subscript or superscript. be consistent!}
%\subsection{The Forward Diffusion Process}
%The forward diffusion process is a Markov chain that gradually corrupts data $\boldsymbol{x}_0$ until it approaches Gaussian noise $\boldsymbol{x}_{T}$. The forward process yields the present state $\boldsymbol{x}_{t}$ from the previous state $\boldsymbol{x}_{t-1}$:
%\begin{equation}
%\begin{split}
%& q(\mathbf{x}_{t}|\mathbf{x}_{t-1})=\mathcal{N}(\mathbf{x}_{t};\sqrt{1-\beta_{t}}\mathbf{x}_{t-1},\beta_{t}\mathbf{I}),
%\end{split}
%\end{equation}
%where $t$ denotes as diffusion step, $\beta_{t}$ is the predefined scale factor.

\subsection{The Reverse Diffusion Process}
The Reverse diffusion Process is a Markov chain that denoises a sampled Gaussian noise to a clean image step by step. Starting from noise $x_T \sim \mathcal{N}(0, \boldsymbol{I})$, the reverse process from latent $\boldsymbol{x}_T$ to clean data $\boldsymbol{x}_0$ is defined as:
\begin{equation}
\begin{split}
& p_{\boldsymbol{\theta}}\left(\boldsymbol{x}_{t-1} \mid \boldsymbol{x}_t\right)=\mathcal{N}\left(\boldsymbol{x}_{t-1} ; \boldsymbol{\mu}_{\boldsymbol{\theta}}\left(\boldsymbol{x}_t, t\right), \Sigma_{\theta} \boldsymbol{I}\right)
\end{split}
\end{equation}
The mean $\boldsymbol{\mu}_{\boldsymbol{\theta}}\left(\boldsymbol{x}_t, t\right)$ is the target we want to estimate by a neural network $\boldsymbol{\theta}$. The variance $\Sigma_{\theta}$ can be either time-dependent constants \cite{ho2020denoising} or learnable parameters \cite{nichol2021improved}.
The reverse process yields the previous state $\boldsymbol{x}_{t-1}$ from the current state $\boldsymbol{x}_{t}$:
\begin{equation}
    \boldsymbol{x}_{t-1} = \frac{1}{\sqrt{\alpha_t}}\left(\boldsymbol{x}_t-\frac{\beta_t}{\sqrt{1-\bar{\alpha}_t}} \boldsymbol{\epsilon}_\theta\left(\boldsymbol{x}_t, t\right)\right) + \sqrt{\Sigma_{\theta}}\boldsymbol{z}
\label{eq5}
\end{equation}
where $\boldsymbol{z} \sim \mathcal{N}(0, \boldsymbol{I})$.
%\vspace{-0.2cm}
In practice, $\boldsymbol{\Tilde{x}}_0$ is usually predicted from $\boldsymbol{x}_t$, then  $\boldsymbol{x}_{t-1}$ is sampled using both $\boldsymbol{\Tilde{x}}_0$ and $\boldsymbol{x}_t$, where $\boldsymbol{\Tilde{x}}_0$ is computed as:
\begin{equation}
\!\!\!\!  \boldsymbol{\tilde{x}}_{0} =  \frac{\boldsymbol{x}_{t}}{\sqrt{\bar{\alpha}_{t}}}-\frac{\sqrt{1-\bar{\alpha}_{t}} \epsilon_{\theta}\left(\boldsymbol{x}_{t}, t\right)}{\sqrt{\bar{\alpha}_{t}}}
\label{eq6}
\end{equation}
 and $\alpha_t=1-\beta_t$ and $\bar{\alpha}_t=\prod_{i=1}^t \alpha_i$.

\section{Method}

\subsection{Overall Framework}

Given a degraded video with $N$ frames $\{I_i\}_{i=0}^N$, our goal is to restore or enhance it to a normal-light clean video $\{\bar{I}_i\}_{i=0}^N$. Our method leverages a pre-trained image diffusion model \cite{dhariwal2021diffusion} for zero-shot video restoration and enhancement. The work in \cite{dhariwal2021diffusion} utilizes a U-Net constructed by layers of 2D convolutional residual blocks and spatial self-attention blocks. We replace all $3\times3$ 2D convolutions with inflated $1\times3\times3$ 3D convolutions to process video. For better temporal consistency, we propose SLR temporal attention, temporal consistency guidance, spatial-temporal noise sharing, and early stopping sampling strategy, the framework is illustrated in Fig. \ref{fig:framework}. It's worth noting that our method is a plug-and-play module, meaning it can be easily incorporated into any diffusion-based image restoration or enhancement method.

\subsection{SLR Temporal Attention}

\begin{figure*}
    \centering
    \includegraphics[width=1.0\linewidth]{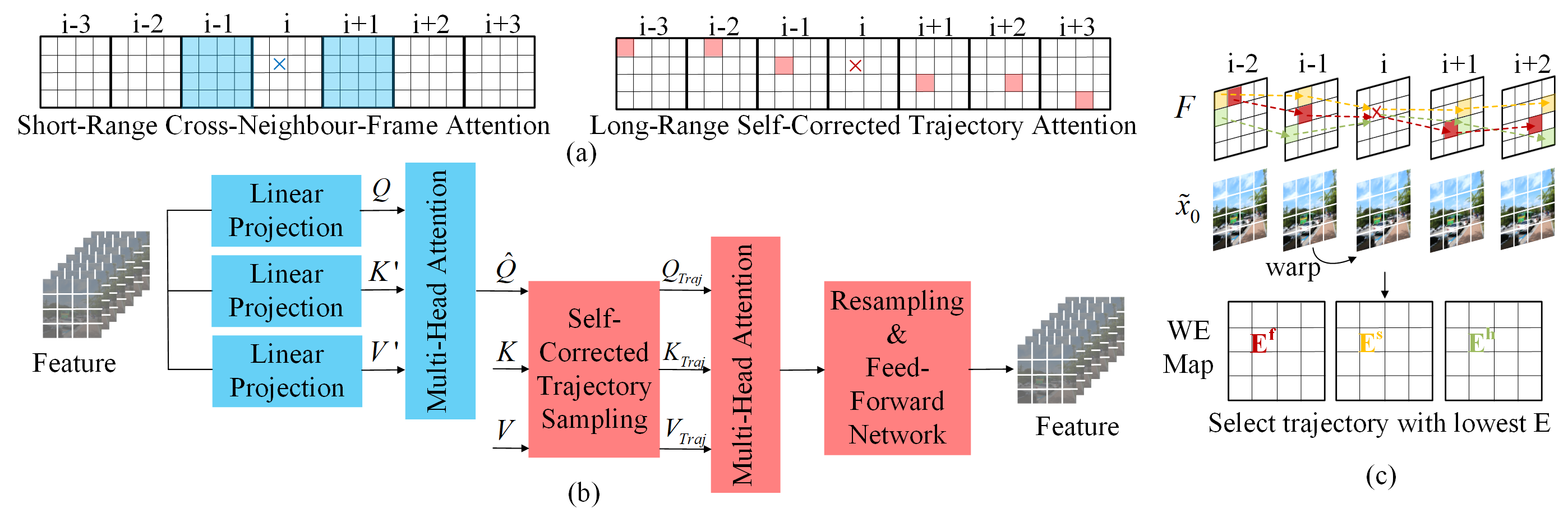}
    \caption{Architecture of the proposed SLR Temporal Attention. (a) The two modules in SLR temporal attention: cross-neighbor-frame attention and self-corrected trajectory attention, focus on short-range and long-range temporal correlation between frames, respectively. (b) The cross-neighbor-frame attention is applied first, and its output serves as the query for the self-corrected trajectory attention. (c) The procedure of self-corrected trajectory sampling. The red, yellow, and green trajectories denote the flow-based, similarity-based, and historically-best trajectories, respectively.}
    \label{fig:SLRTA}
\end{figure*}

We propose SLR temporal attention to strengthen the temporal consistency of video restoration and enhancement results. Since the decoder layers are less noisy than the encoder layer in the sampling, we replace spatial self-attention layers in the U-Net decoder with our SLR temporal attention layer, which consists of two modules: cross-neighbor-frame attention and self-corrected trajectory attention, as shown in Fig. \ref{fig:SLRTA}. 

\subsubsection{Cross-Neighbor-Frame Attention} 

For the spatial self-attention layer, the query, key, and value $Q$, $K$, $V$ are obtained by linear projection of the feature $v_i$ from $I_i$, the corresponding self-attention output is produced by
 $\textit{Self\_Attn}(Q,K,V)=\textit{Softmax}(\frac{QK^T}{\sqrt{d}})\cdot V$ with
\begin{equation}
  Q=W^Qv_i, K=W^Kv_i, V=W^Vv_i,
\end{equation}
where $W^Q$, $W^K$, $W^V$ are pre-trained matrices that project the inputs to $Q$, $K$, $V$ respectively. Cross-frame attention uses the key $K'$ and value $V'$ from other frames, which has been widely used for zero-shot video editing. For video editing, besides the previous frame, the first frame is also used to maintain global coherence in terms of generated content. However, for video restoration, we find that the bidirectional neighbor frames are more suitable for maintaining the temporal consistency. The cross-neighbor-frame attention output is produced by
$\textit{CrossNeighFrame\_Attn}(Q,K',V')=\textit{Softmax}(\frac{QK'^T}{\sqrt{d}})\cdot V'$ with
\begin{equation}
  Q=W^Qv_i, K'=W^K(v_{i-1}||v_{i+1}), V'=W^V(v_{i-1}||v_{i+1}),
\end{equation}
where $||$ represents concatenation. 
To reduce the computation cost, we only leverage the previous and next neighbor frame to capture the short-range temporal correlation between frames, and utilize the following self-corrected trajectory attention to capture the long-range temporal correlation. As shown in Fig. \ref{fig:SLRTA}, the colored $\times$ denotes the position of query in attention, and the colored square denotes the position of key and value. The output of the cross-neighbor-frame serves as the query $\hat{Q}$ for self-corrected trajectory attention.

\subsubsection{Self-Corrected Trajectory Attention} 
\label{sec:ta}

To further improve the temporal consistency, optical flow-guided attention \cite{cong2023flatten} was proposed, which samples patch trajectories according to optical flow and performs the attention on the patch embeddings in the same trajectory. However, inaccurate flow-based trajectories from inaccurate optical flow limit the performance. Especially for video restoration and enhancement, the input degraded frames damage the optical flow calculation. Performing restoration twice and calculating the optical flow after the first restoration without flow-guided attention results in a doubling of the inference time, and the gap between two restoration processes also influences the suitability of the optical flow. In view of this, we propose a self-corrected strategy to progressively correct the trajectories in the sampling process. For the U-Net in \cite{dhariwal2021diffusion}, the original spatial self-attention is applied to the features with resolution $32\times32$, $16\times16$, and $8\times8$. We inject our self-corrected trajectory attention to the largest resolution $32\times32$. In the $t$ step of sampling, for each pixel in the feature map, its flow-based trajectory can be calculated from the downsampled optical flow. The optical flow is calculated on the $\boldsymbol{\tilde{x}}_0$ of $t+1$ step. The clean image $\boldsymbol{\tilde{x}}_0$ can be directly inferred when given $\boldsymbol{x}_t$ by the Eq. \ref{eq6} in every timestep $t$. As $t$ decreases, $\boldsymbol{\tilde{x}}_0$ will have better quality, resulting in more accurate optical flow. In addition, we propose the similarity-based trajectory and historically-best trajectory to correct the flow-based trajectory. 

Given the diffusion feature $F_i$ and $F_{i-1}$ of frame $I_i$ and $I_{i-1}$, the cosine similarity between pixel pairs $(p,q)$ in the feature can be formulated as
%\begin{equation}
%  S=\frac{F_i\cdot F_{i-1}}{||F_i||\,||F_{i-1}||}.
%\end{equation}
\begin{equation}
  S(p,q)=\frac{F_i(p)\cdot F_{i-1}(q)}{||F_i(p)||\,||F_{i-1}(q)||}.
\end{equation}
The similarity-based trajectory between frame $I_i$ and $I_{i-1}$ can be obtained from the pixel pairs with the highest similarity. We define the best trajectory for a pixel in the feature as the trajectory 
that can achieve the best temporal consistency on the corresponding patch of $\boldsymbol{\tilde{x}}_0$. For every timestep $t$, the historically-best trajectory is defined as the best trajectory at step $t+1$. The inaccurate flow-based trajectory can be compensated for by similarity-based trajectory and historically-best trajectory. Specifically, for each pixel in feature $F_i$ at step $t$, we compute $\boldsymbol{\tilde{x}}_0$ of three different trajectories, respectively. Each pixel corresponds to a $8\times8$ patch in $\boldsymbol{\tilde{x}}_0$, we warp the previous frame and compute the average warp error of this patch area. The trajectory with the lowest warp error serves as the final trajectory of this pixel at step $t$, i.e., the best trajectory at step $t$ and the historically-best trajectory for the step $t-1$. This procedure is shown in Fig. \ref{fig:SLRTA} (c). The trajectory attention output is produced by
$\text{Traj\_Attn}(Q_{Traj},K_{Traj},V_{Traj})=\text{Softmax}(\frac{Q_{Traj}K_{Traj}^T}{\sqrt{d}})\cdot V_{Traj}$ with
\begin{equation}
  Q_{Traj}=\hat{Q}[p], K_{Traj}=K[Traj-p], V_{Traj}=V[Traj-p].
\end{equation}
where [p] denotes the sampling value of pixel $p$, and [Traj-p] denotes the sampling values of pixels in the trajectory of $p$ except for $p$ itself.
We find that the $\boldsymbol{\tilde{x}}_0$ is not always clean in the whole sampling process.
At the beginning of sampling, $\boldsymbol{\tilde{x}}_0$ has a lower signal-to-noise ratio, where the image contents are unrecognizable and have a lot of noise. In the middle part of sampling, $\boldsymbol{\tilde{x}}_0$ has smooth content which cannot be used to compute precise optical flow. Only in the second half of sampling does the diffusion model slowly generate rich content and details, which are suitable for computing precise optical flow. In practice, we only apply self-corrected trajectory attention after the current diffusion step $t\textless T_{TA}$, where $T_{TA}$ is set to 100 for the GDP backbone. We utilize RAFT \cite{teed2020raft,jeong2023ground,cong2023flatten} to calculate optical flow, and utilize forward-backward consistency check to generate occlusion mask for warp error \cite{lai2018learning}.

\subsection{Temporal Consistency Guidance}

Since the attention module only exists on the features with $32\times32$ and lower resolution, the higher resolution ($64\times64$, $128\times128$, $256\times256$) features can also cause the temporal inconsistency of output. Therefore we propose the temporal consistency guidance to directly constrain the final output. The temporal consistency guidance is categorized into pixel-level consistency and semantic-level consistency. For pixel-level consistency, we compute the optical flow and occlusion mask between the $\boldsymbol{\tilde{x}}_0$ of frame $I_i$ and $I_{i-1}$ (denoted by $\boldsymbol{\tilde{x}}_0^{i}$ and $\boldsymbol{\tilde{x}}_0^{i-1}$) in the step $t+1$, then constrain $\boldsymbol{\tilde{x}}_0^{i}$ and $\boldsymbol{\tilde{x}}_0^{i-1}$ in step $t$ with pixel-level consistency
\begin{equation}
\mathcal{L}^{PC}_{\boldsymbol{\tilde{x}}_{0}} = \sum_{i=0}^{N} M_{i} \left\| \boldsymbol{\tilde{x}}_0^{i} - warp(\boldsymbol{\tilde{x}}_0^{i-1}, f_{i}) \right\|_1
\end{equation}
where $M_{i}$ is the predicted occlusion mask, $f_{i}$ is the predicted optical flow. For semantic-level consistency, the neighbour frames should have similar sematic information. We utilize the image encoder of CLIP to extract the embedding $E^{i}$ and $E^{i-1}$ of $\boldsymbol{\tilde{x}}_0^{i}$ and $\boldsymbol{\tilde{x}}_0^{i-1}$, the semantic-level consistency can be formulated as 
\begin{equation}
\mathcal{L}^{SC}_{\boldsymbol{\tilde{x}}_{0}} = 1-\frac{E_i\cdot E_{i-1}}{||E_i||\,||E_{i-1}||}.
\end{equation}
the totally temporal consistency can be formulated as 
\begin{equation}
\mathcal{L}^{TC}_{\boldsymbol{\tilde{x}}_{0}} = \mathcal{L}^{PC}_{\boldsymbol{\tilde{x}}_{0}} + \gamma \mathcal{L}^{SC}_{\boldsymbol{\tilde{x}}_{0}}
\end{equation}
Then we apply gradient guidance \cite{fei2023generative} to guide the sampling process. Specifically, we sample $\boldsymbol{x}_{t-1}$ by $\mathcal{N}\left(\mu+s\nabla_{\boldsymbol{\tilde{x}}_{0}} \mathcal{L}^{TC}_{\boldsymbol{\tilde{x}}_{0}}, \sigma^2\right)$, $s$ is gradient scale. Since only in the second half of sampling, $\boldsymbol{\tilde{x}}_0$ is suitable to compute optical flow, we apply pixel-level consistency guidance after the current diffusion step $t\textless T_{TC}$, which is set to 300 for the GDP backbone. We apply semantic-level consistency guidance throughout the entire sampling process, compensating for the absence of pixel-level consistency guidance in the early steps.

\subsection{Spatial-Temporal Noise Sharing}

Recently, \cite{chen2024deconstructing} demonstrates that the denoising process plays an important role in the denoising diffusion model. Actually, the noise in the sampling process controls the final generated color and details. For the same degraded frame, different noise $\boldsymbol{x}_{T}$ and $\boldsymbol{z}$ in the reverse diffusion process will lead to different colors and details in the result. For better temporal consistency, we propose to share the same $\boldsymbol{x}_{T}$ and $\boldsymbol{z}$ in Eq. \ref{eq5} between all frames, which encourages the diffusion model to generate the same details in the static areas. We used the predicted optical flow and occlusion mask to blend the $\boldsymbol{z}$ of degraded frame $I_i$ and $I_{i-1}$, which are denoted by the $\boldsymbol{z}^{i}$ and $\boldsymbol{z}^{i-1}$. We propose to blend $\boldsymbol{z}$ rather than to blend $\boldsymbol{\tilde{x}}_0$, $\boldsymbol{\tilde{x}}_t$ or U-Net feature since the latter usually leads to motion ghost and unpleasant artifacts. The blending process can be formulated as
\begin{equation}
\boldsymbol{z}^{i} = M_{i}(\lambda\boldsymbol{z}^{i}+(1-\lambda)\boldsymbol{z}^{i-1})+(1-M_{i})\boldsymbol{z}^{i}
\end{equation}
The blending process shares noise between the corresponding pixels in different frames, which encourages the diffusion model to generate the same details in these dynamic areas. 

\subsection{Early Stopping Sampling Strategy}

In the above sections, we find that $\boldsymbol{x}_0$ firstly reconstructs the low-frequency component of the frame, then reconstructs the high-frequency component in the sampling, the temporal flicker easily increases at the end of the reverse diffusion process. Besides, the real-world degraded images often suffer from noise. When enhancing low-light videos, the diffusion model reconstructs the high-frequency noise at the end of sampling, thereby reducing temporal consistency. We propose an early stopping sampling strategy, which stops sampling after $T_{ES}$, preventing $\boldsymbol{x}_0$ from reconstructing noise or unconsistent high-frequency details. We take the early stopping $\boldsymbol{x}_0$ as the final result.

\begin{table}[t]
\centering
%\begin{tabular}{m{5cm}m{0.7cm}<{\centering}m{0.7cm}<{\centering}}
%\begin{tabular}{m{2.2cm}m{1.0cm}<{\centering}m{0.9cm}<{\centering}m{0.9cm}<{\centering}m{0.9cm}<{\centering}m{0.9cm}<{\centering}m{0.9cm}<{\centering}m{0.9cm}<{\centering}}
\resizebox{0.49\textwidth}{13.5mm}{
\addtolength{\tabcolsep}{-5pt}
\begin{tabular}{l|c|c|c|c|c|c}
%\begin{tabular}{m{5cm}m{0.7cm}m{0.7cm} }
%\begin{tabular}
\toprule
Methods                                & PSNR$\uparrow$  & SSIM$\uparrow$  & FID$\downarrow$ & WE$\downarrow$  & FS$\uparrow$ & OFME$\downarrow$\\
\hline
VRT                                       &\textbf{23.68}       &\textbf{0.7434}      &157.87                &\underline{0.4797}   &\underline{0.9858}    &\textbf{0.1563}                \\
DDNM                                       &23.46                &0.6876               &\underline{110.13}    &1.3103               &0.9513                &0.3212                  \\
DDNM+ZVRD                                  &\underline{23.53}    &\underline{0.6925}   &\textbf{106.84}       &0.5339               &0.9754                &\underline{0.2596}                    \\
GDP                                        &20.44                &0.5252               &171.59                &4.0327               &0.8950                &4.3595                  \\
GDP+ZVRD                                   &21.39                &0.5843               &167.44                &\textbf{0.4234}      &\textbf{0.9885}       &0.9948                   \\
\bottomrule
\end{tabular}
}
\caption{Quantitative comparison with state-of-the-art methods for 4$\times$ video super-resolution. The best results are highlighted in bold and the second best results are underlined. WE is expressed as a percentage (\%). VRT is a supervised method, the others are zero-shot methods.}
\label{ComparisonVSR}
\end{table}

\begin{table}[t]
\centering
%\begin{tabular}{m{5cm}m{0.7cm}<{\centering}m{0.7cm}<{\centering}}
%\begin{tabular}{m{3.0cm}m{2.5cm}<{\centering}m{1.6cm}<{\centering}m{1.3cm}<{\centering}m{1.3cm}<{\centering}m{1.8cm}}
\resizebox{0.49\textwidth}{13.5mm}{
\addtolength{\tabcolsep}{-5pt}
\begin{tabular}{l|c|c|c|c|c|c}
%\begin{tabular}{m{5cm}m{0.7cm}m{0.7cm} }
%\begin{tabular}
\toprule
Methods                                      & PSNR$\uparrow$  & SSIM$\uparrow$  & FID$\downarrow$ & WE$\downarrow$  & FS$\uparrow$ & OFME$\downarrow$\\
\hline
BiSTNet                                        &23.67             &0.9920             &131.04             &\underline{1.1752}      &0.9787                   &\textbf{0.1213} \\
DDNM                                            &24.60             &\underline{0.9932} &\underline{123.29} &2.4315              &0.9371                    &0.7094 \\
DDNM+ZVRD                                      &\textbf{24.86}    &\textbf{0.9945}    &\textbf{121.87}    &1.2992              &\textbf{0.9866}           &\underline{0.1386} \\
GDP                                            &24.58             &0.9333             &134.56             &1.3125              &0.9176                    &0.3658 \\
GDP+ZVRD                                        &\underline{24.64} &0.9416             &133.39             &\textbf{0.9208}     &\underline{0.9850}        &0.2875 \\
\bottomrule
\end{tabular}
}
\caption{Quantitative comparison with state-of-the-art methods for video colorization. The best results are highlighted in bold and the second best results are underlined. WE is expressed as a percentage (\%). BiSTNet is a supervised method, the others are zero-shot methods.}
\label{ComparisonVColor}
\end{table}

\begin{table}[t]
\centering
%\begin{tabular}{m{5cm}m{0.7cm}<{\centering}m{0.7cm}<{\centering}}
%\begin{tabular}{m{3.0cm}m{2.5cm}<{\centering}m{1.6cm}<{\centering}m{1.3cm}<{\centering}m{1.3cm}<{\centering}m{1.8cm}}
\resizebox{0.49\textwidth}{12mm}{
\addtolength{\tabcolsep}{-5pt}
\begin{tabular}{l|c|c|c|c|c|c}
%\begin{tabular}{m{5cm}m{0.7cm}m{0.7cm} }
%\begin{tabular}
\toprule
Methods                                       & PSNR$\uparrow$  & SSIM$\uparrow$  & FID$\downarrow$ & WE$\downarrow$  & FS$\uparrow$ & OFME$\downarrow$\\
\hline
FastLLVE                                       &12.76             &0.6572              &261.69            &0.7236               &0.9791               &0.5306   \\
SGZ                                             &17.22             &0.6576              &\textbf{49.49}    &\underline{0.4548}   &\underline{0.9904}   &0.3844 \\
GDP                                             &\underline{17.35} &\underline{0.8072}  &62.05             &0.6029               &0.9827               &\underline{0.3533} \\
GDP+ZVRD                                        &\textbf{17.56}    &\textbf{0.8237}     &\underline{60.54} &\textbf{0.3352}      &\textbf{0.9910}      &\textbf{0.3181}  \\
\bottomrule
\end{tabular}
}
\caption{Quantitative comparison with state-of-the-art methods for low-light video enhancement. The best results are highlighted in bold and the second best results are underlined. WE is expressed as a percentage (\%). FastLLVE is a supervised method, the others are zero-shot methods.}
\label{ComparisonVE}
\end{table}

\begin{table}[t]
\centering
%\begin{tabular}{m{5cm}m{0.7cm}<{\centering}m{0.7cm}<{\centering}}
%\begin{tabular}{m{3.0cm}m{2.5cm}<{\centering}m{1.6cm}<{\centering}m{1.3cm}<{\centering}m{1.3cm}<{\centering}m{1.8cm}}
\resizebox{0.49\textwidth}{10mm}{
\addtolength{\tabcolsep}{-5pt}
\begin{tabular}{l|c|c|c|c|c|c}
%\begin{tabular}{m{5cm}m{0.7cm}m{0.7cm} }
%\begin{tabular}
\toprule
Methods                          & PSNR$\uparrow$  & SSIM$\uparrow$  & FID$\downarrow$ & WE$\downarrow$  & FS$\uparrow$ & OFME$\downarrow$\\
\hline
DiffBIR                          &24.47	                 &\underline{0.6727}	 &32.03	              &0.6907	            &\underline{0.9825}	 &0.1328   \\
DiffIR2VR                   &\underline{24.49}	      &0.6718	         &\underline{30.41}	  &\underline{0.6818}	&0.9820	             &\underline{0.1312}     \\
DiffBIR+ZVRD                     &\textbf{24.55}	      &\textbf{0.6799}	 &\textbf{30.12}	  &\textbf{0.4923}	    &\textbf{0.9898}	 &\textbf{0.1145}     \\
\bottomrule
\end{tabular}
}
\caption{Quantitative comparison with state-of-the-art methods for 4$\times$ blind video super-resolution on the DAVIS dataset. The best results are highlighted in bold and the second best results are underlined. WE is expressed as a percentage (\%).}
\label{ComparisonVBSR}
\end{table}

\section{Experiments}

\subsection{Test Datasets}

For video super-resolution, we collected 18 gt videos from commonly used test datasets REDS4 \cite{nah2019ntire}, Vid4 \cite{liu2013bayesian}, and UDM10 \cite{yi2019progressive}. For video deblurring, we collected 10 ground truth (GT) videos from the dataset REDS \cite{nah2019ntire}. For video denoising, we collected 15 GT videos from the commonly used test dataset Set8 \cite{tassano2020fastdvdnet} and DAVIS \cite{pont20172017}. For video inpainting, we collected 20 GT videos from the commonly used DAVIS \cite{pont20172017} dataset. For video colorization, we use the GT videos from the Videvo20 \cite{lai2018learning} dataset, which is one of the mainly used datasets for video colorization.  We follow \cite{chung2022diffusion,wang2022zero,fei2023generative} to apply linear degradation to GT videos to construct corresponding degraded videos for video super-resolution, deblurring, denoising, inpainting, and colorization respectively. For low-light video enhancement, we collected 10 paired low-normal videos from the DID dataset \cite{fu2023dancing} which was captured in the real world. Due to the slow sampling speed of DDPM and a test video containing a lot of frames, we first center crop the frames along the shorter edge and then resize them to 256$\times$256, which matches the image size of the diffusion model. Our method can be combined with a patch-based strategy in \cite{fei2023generative} to process any-size videos.

%For each low and normal video, We center-crop the frames along the shorter edge, and then resize them to 256$\times$256.

\begin{figure*}
    \centering
    \includegraphics[width=0.7\linewidth]{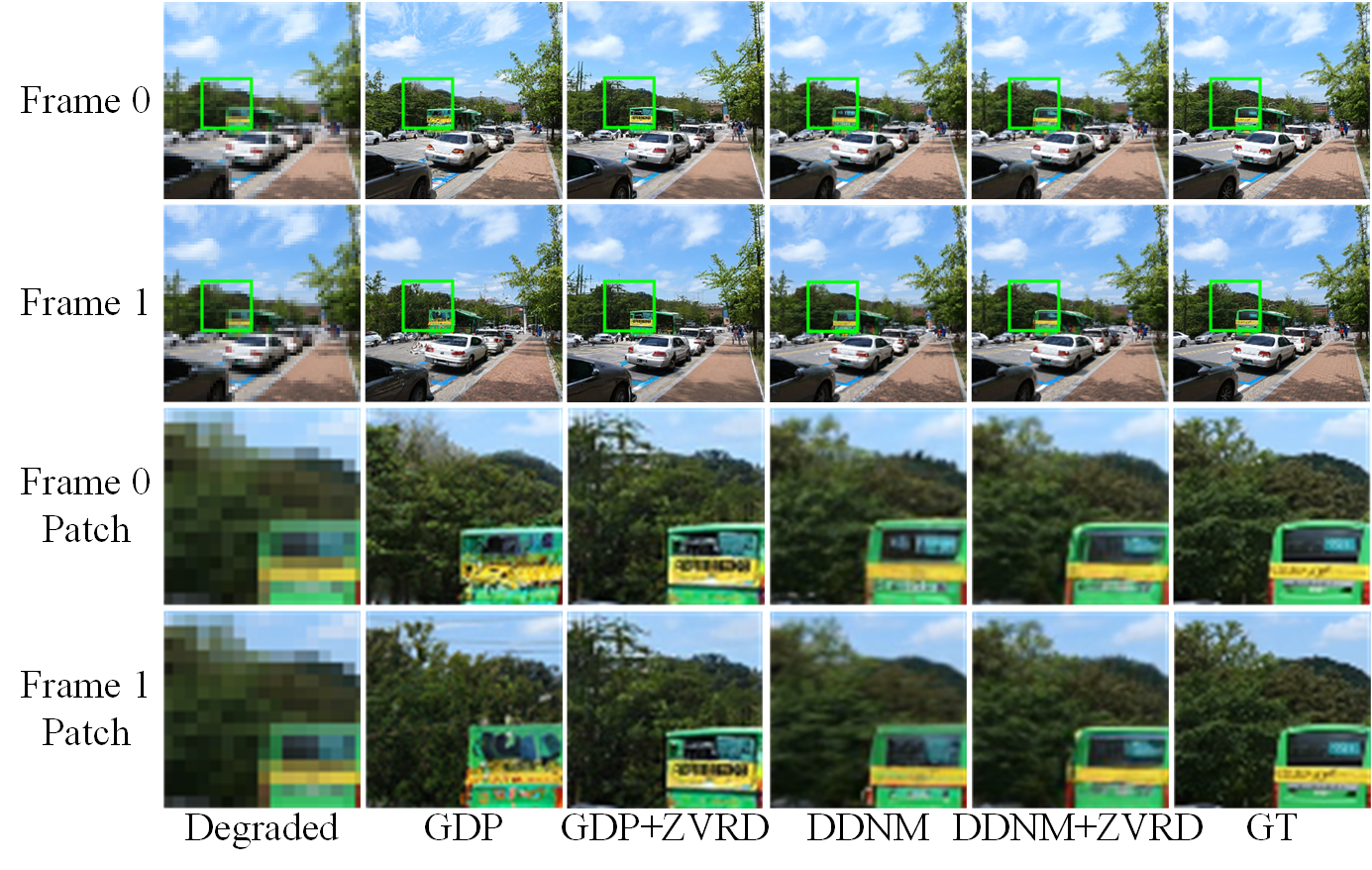}
    \caption{Visual quality comparison for video super-resolution. Zoom in for better observation.}
    \label{fig:videosr}
\end{figure*}

\subsection{Comparison with State-of-the-art Methods}
\label{sec:comparison}

We utilize six metrics to evaluate the restoration and enhancement quality. Besides the commonly used metrics PSNR, SSIM, and FID, we utilize Warping Error (WE) \cite{lai2018learning}, Frame Similarity (FS) \cite{wu2023tune,chen2023control,qi2023fatezero}, and optical flow map error (OFME) \cite{wang2024videocomposer,chen2023control} to evaluate temporal consistency. In our supplementary file, we also provide the user study for temporal consistency evaluation. FS was introduced to assess semantic consistency between generated frames by calculating the similarities of CLIP embeddings of output video frames. OFME was introduced to measure the movement consistency in video synthesis and editing. We extend it to evaluate video restoration and enhancement by calculating the optical flow map error between restored/enhanced frames and ground truth frames. Since our method is a plug-and-play method, we choose three state-of-the-art zero-shot image restoration methods, namely DPS \cite{chung2022diffusion}, DDNM \cite{wang2022zero} and GDP \cite{fei2023generative} as our compared methods and backbones. We utilize their content constraints in our method and extend them for zero-shot video restoration, respectively. Besides the three backbones, we also compare with VRT (supervised training) \cite{liang2024vrt} for video super-resolution and deblurring, FastDVDNet (supervised training) \cite{tassano2020fastdvdnet} and UDVD (unsupervised training) \cite{sheth2021unsupervised} for video denoising. For video inpainting, we compare with zero-shot image inpainting method RePaint \cite{lugmayr2022repaint}. For video colorization, we compare with the supervised method BiSTNet \cite{yang2024bistnet}. For low-light video enhancement, we compared with the supervised method FastLLVE \cite{li2023fastllve} and zero-shot video enhancement method SGZ \cite{zheng2022semantic}. 

%Table \ref{ComparisonVSR}, \ref{ComparisonVDB}, \ref{ComparisonVDN}, \ref{ComparisonVInp}, \ref{ComparisonVColor}, \ref{ComparisonVE} 
Tables \ref{ComparisonVSR}-\ref{ComparisonVE} list the quantitative results for the video super-resolution, video colorization, and low-light video enhancement, respectively. It can be observed that by inserting our method in existing zero-shot image restoration methods (DDNM+ZVRD, GDP+ZVRD, DPS+ZVRD), the temporal consistency can be obviously improved. For 4$\times$ video super-resolution, on the basis of DDNM, the WE is decreased to nearly 1/3 of the original, and the FID is increased and outperforms the supervised method VRT. On the basis of GDP, the WE is decreased to about 1/10 of the original, which is better than VRT. Our method achieves nearly 1 dB gain for PSNR. For all tasks, our method can improve the performance in most of the six metrics. For video colorization, our method can boost DDNM to outperform the supervised method BiSTNet in four metrics. For low-light video enhancement, our method can enhance GDP in five metrics, surpassing other methods. It demonstrates the effectiveness of our method. Due to the page limit, we give the quantitative results for the video deblurring, video denoising, and video inpainting in the supplementary material.

Figs. \ref{fig:videosr}, \ref{fig:videoenhance} present the visual comparison results on the evaluation data for video super-resolution and low-light video enhancement, respectively. Due to the page limit, visual comparison results on more tasks are shown in the supplementary file. Fig. \ref{fig:videosr} presents the results of the four methods on the first and second frames of the video. For GDP, the details of the tree and bus are not consistent on the two frames, and the shape of the car is also obviously different. For DDNM, there are different contents on the window of the bus. Our method (DDNM+ZVRD, GDP+ZVRD) can restore temporal consistent results on both tree and bus. As shown in Fig. \ref{fig:videoenhance}, GDP has different global light and different details on the table. Our method has better temporal consistency on global light and local details. 

%Due to the page limit, we give a user study in the supplementary materials.

Besides the above linear restoration tasks and non-linear, blind enhancement task, our method can also be applied to blind restoration tasks with complex real-world degradation. Following the settings of DiffIR2VR \cite{yeh2024diffir2vr}, we use DiffBIR as the backbone for blind video super-resolution and evaluate on DAVIS testing sets. Low-quality videos are generated using the degradation pipeline of RealBasicVSR. Our method achieves the best performance for all six metrics as shown in Table \ref{ComparisonVBSR}. Since \cite{yeh2024diffir2vr} relies on optical flow which is inaccurate and directly merges similar tokens of attention blocks between frames, they tend to generate blurry results. Our SLR temporal attention provides a softer way to solve the issue of temporal consistency. The self-corrected trajectory attention in SLR temporal attention can adaptly compensate for inaccurate optical flow through similarity-based trajectory and historically-best trajectory. Thus our method generates sharper results.

\begin{figure}
    \centering
    \includegraphics[width=0.9\linewidth]{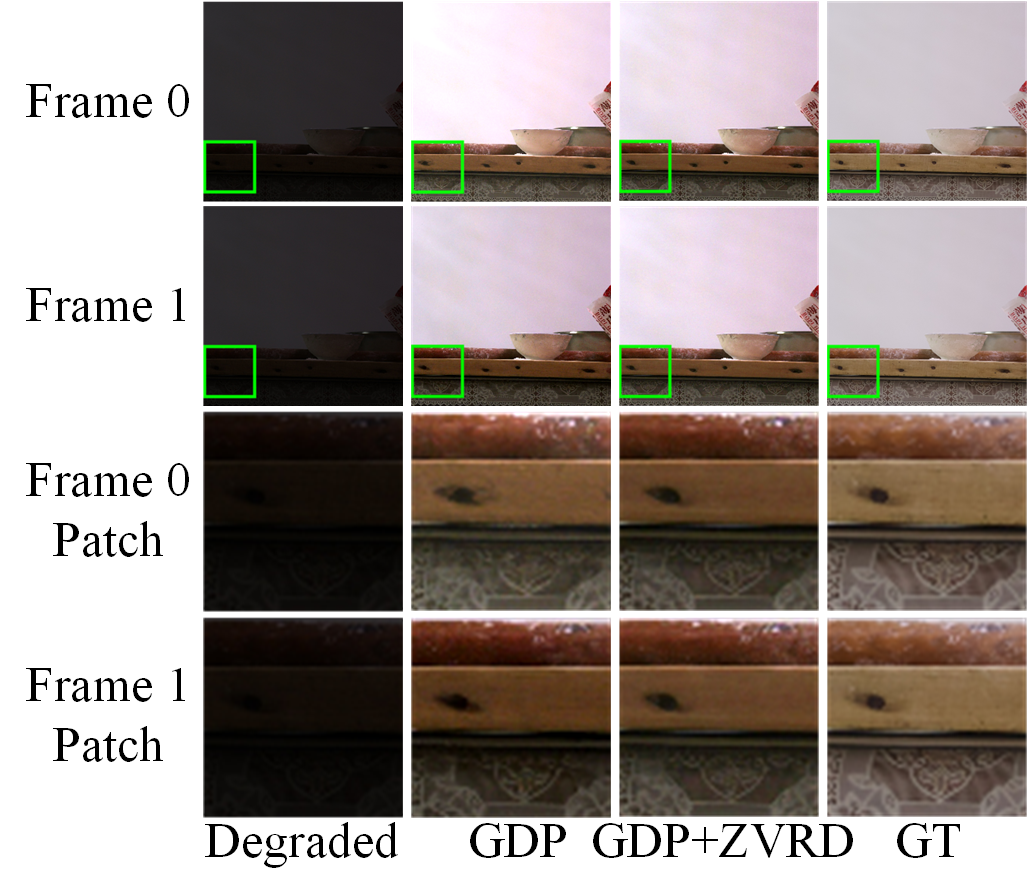}
    \caption{Visual quality comparison for low-light video enhancement. Zoom in for better observation.}
    \label{fig:videoenhance}
\end{figure}

\subsection{Ablation Study}

\begin{table}[t]
\centering
\resizebox{0.48\textwidth}{21mm}{
\addtolength{\tabcolsep}{-1pt}
\begin{tabular}{cccccc}
\toprule
SLTA                      & $\times$   & $\checkmark$ & $\checkmark$     & $\checkmark$    & $\checkmark$    \\\hline                          
TCG                       & $\times$   & $\times$     & $\checkmark$     & $\checkmark$    & $\checkmark$    \\\hline
STNS                      & $\times$   & $\times$     & $\times$         & $\checkmark$    & $\checkmark$     \\\hline
ESSS                      & $\times$   & $\times$     & $\times$         & $\times$        & $\checkmark$     \\ \hline
PSNR$\uparrow$            & 20.44      & 21.12        & 21.15            & 21.42           & 21.39 \\
SSIM$\uparrow$            & 0.5252     & 0.5611       & 0.5623           & 0.5847          & 0.5843 \\
FID$\downarrow$           & 171.59     & 168.89       & 168.62           & 167.35          & 167.44 \\
WE$\downarrow$            & 4.0327     & 2.2806       & 1.9281           & 0.4586          & 0.4234 \\
FS$\uparrow$              & 0.8950     & 0.9220       & 0.9653           & 0.9867          & 0.9885 \\
OFME$\downarrow$          & 4.3595     & 1.8654       & 1.3540           & 0.9972          & 0.9948 \\
\bottomrule
\end{tabular}
}
\caption{Ablation study for SLR Temporal Attention (SLTA), temporal consistency guidance (TCG), spatial-temporal noise sharing (STNS) and early stopping sampling strategy (ESSS) on 4$\times$ video super-resolution task. WE is expressed as a percentage (\%).}
\label{Ablation}
\end{table}

In this section, we perform an ablation study to demonstrate the effectiveness of the proposed SLR Temporal Attention, Temporal Consistency Guidance, Spatial-Temporal Noise Sharing, and Early Stopping Sampling Strategy. Take video super-resolution as an example, Table \ref{Ablation} lists the quantitative comparison results on evaluation data by adding these modules one by one. It can be observed that SLR Temporal Attention can bring 0.68 dB gain for PSNR, 2.7 gain for FID, nearly 1.75 gain for WE, and nearly 2.5 gain for OFME. When adding Temporal Consistency Guidance, WE is decreased by nearly 0.35, and FS is increased by 0.0433. Spatial-Temporal Noise Sharing can bring 0.27 dB gain for PSNR and reduce WE by nearly 1.5. It is ranked the second in terms of gain for the metric WE. Early Stop Sampling Strategy can further reduce the WE, and OFME and improve FS while keeping other metrics basically unchanged. Due to the page limit, we give a more detailed ablation study in the supplementary material.

\section{Conclusion}
In this paper, we propose the first framework for zero-shot video restoration and enhancement which uses a pretrained image diffusion model and is training-free. By replacing the spatial self-attention layer with the proposed SLR temporal attention layer, the pre-trained image diffusion model can utilize the temporal correlation between frames. To further strengthen the temporal consistency of results, we propose temporal consistency guidance, spatial-temporal noise sharing, and an early stopping sampling strategy. Experimental results demonstrate the superiority of the proposed method.

\section{Acknowledgements}
This work was supported in part by National Natural Science Foundation of China under Grant 62472308, 62231018 and 62171309.

\bibliography{aaai25}

\end{document}

\maketitle

This supplementary file provides details which were not presented in the main paper due to page limitations. In the following, we first give the detailed experiment settings. Then we present more ablation study and comparison results. Finally, a demo for video results comparison is given.

\section{Experiment Settings}

Algo. \ref{algo1} shows our sampling process.

\begin{algorithm}[h]\footnotesize
	\caption{\textbf{Sampling process}: Given a diffusion model $\left(\mu_{\theta}\left(\boldsymbol{x}_{t}\right), \Sigma_{\theta}\left(\boldsymbol{x}_{t}\right)\right)$, corrupted video $\{I_i\}_{i=0}^N$.}
	\KwIn{Corrupted video $\{I_i\}_{i=0}^N$, gradient scale $s$, content constraint $ccf$, hyper-parameters $T_{ES}$, $\gamma$, and $\lambda$.}
	\KwOut{Output restored or enhanced video $\{I'_i\}_{i=0}^N$}
    Sample $\boldsymbol{x}_{T}^{0}$ from $\mathcal{N}(0, \mathbf{I})$
        
    \For{$i$ from 1 to $N$}{
            
        $\boldsymbol{x}_{T}^{i-1}$ = $\boldsymbol{x}_{T}^{0}$
        
        $\boldsymbol{x}_{T}^{i}$ = $\boldsymbol{x}_{T}^{0}$
        
    	\For{$t$ from $T$ to $T_{ES}$}{
    	    $\mu_{i-1}, \sigma_{i-1}^2 = \mu_{\theta}\left(\boldsymbol{x}_{t}^{i-1}\right), \Sigma_{\theta}\left(\boldsymbol{x}_{t}^{i-1}\right)$
    
            $\mu_{i}, \sigma_{i}^2 = \mu_{\theta}\left(\boldsymbol{x}_{t}^{i}\right), \Sigma_{\theta}\left(\boldsymbol{x}_{t}^{i}\right)$
    	    
    	    $\boldsymbol{\tilde{x}}_{0}^{i-1} =  \frac{\boldsymbol{x}_{t}^{i-1}}{\sqrt{\bar{\alpha}_{t}}}-\frac{\sqrt{1-\bar{\alpha}_{t}} \epsilon_{\theta}\left(\boldsymbol{x}_{t}^{i-1}, t\right)}{\sqrt{\bar{\alpha}_{t}}}$

    	    $\boldsymbol{\tilde{x}}_{0}^{i} =  \frac{\boldsymbol{x}_{t}^{i}}{\sqrt{\bar{\alpha}_{t}}}-\frac{\sqrt{1-\bar{\alpha}_{t}} \epsilon_{\theta}\left(\boldsymbol{x}_{t}^{i}, t\right)}{\sqrt{\bar{\alpha}_{t}}}$
    
            $ccf(\boldsymbol{\tilde{x}}_{0}^{i-1}, I_{i-1})$ 
            
            $ccf(\boldsymbol{\tilde{x}}_{0}^{i}, I_{i})$ 
    
            \eIf{$t\textless T_{TC}$}{

                $ \mathcal{L}^{TC}_{\boldsymbol{\tilde{x}}_{0}^{i}} = \mathcal{L}^{PC}_{\boldsymbol{\tilde{x}}_{0}} + \gamma \mathcal{L}^{SC}_{\boldsymbol{\tilde{x}}_{0}} $
            }
            {
                $ \mathcal{L}^{TC}_{\boldsymbol{\tilde{x}}_{0}^{i}} = 0$
            }
    
            Sample $\boldsymbol{z}_{t}^{0}$ from $\mathcal{N}(0, \mathbf{I})$
            
            $\boldsymbol{z}_{t}^{i-1}$ = $\boldsymbol{z}_{t}^{0}$
            
            $\boldsymbol{z}_{t}^{i}$ = $\boldsymbol{z}_{t}^{0}$
            
            $\boldsymbol{z}_{t}^{i} = M_{i}(\lambda\boldsymbol{z}_{t}^{i}+(1-\lambda)\boldsymbol{z}_{t}^{i-1})+(1-M_{i})\boldsymbol{z}_{t}^{i}$
    
            Sample $\boldsymbol{x}_{t-1}^{i-1}$ by $\boldsymbol{x}_{t-1}^{i-1} = \mu_{i-1} + \sigma_{i-1}^2\boldsymbol{z}_{t}^{i-1}$
    
    	    Sample $\boldsymbol{x}_{t-1}^{i}$ by $\boldsymbol{x}_{t-1}^{i} = \mu_{i} + s\nabla_{\boldsymbol{\tilde{x}}_{0}^{i}} \mathcal{L}^{TC}_{\boldsymbol{\tilde{x}}_{0}^{i}}  + \sigma_{i}^2\boldsymbol{z}_{t}^{i}$
      
    	}
        \Return $\boldsymbol{x}_{0}^{i}$
    }
	
\label{algo1}
\vspace{-0.1em}
\end{algorithm}

For the video restoration tasks, we follow the settings of the linear degradation operator from \cite{chung2022diffusion,wang2022zero,fei2023generative}. For super-resolution with $n$, we set degradation operator as the average-pooling operator $\begin{bmatrix}\frac{1}{n^{2}} & ... & \frac{1}{n^{2}}\end{bmatrix}$ that averages each patch into a single value. For deblurring, the motion blur kernel is 33$\times$33 with a strength of 0.5. For denoising, we add Gaussian noise with $\sigma=50$. For inpainting, the degradation operator is the mask operator. For colorization, the degradation operator is a pixel-wise operator $\begin{bmatrix}\frac{1}{3} & \frac{1}{3} & \frac{1}{3}\end{bmatrix}$ that converts each RGB channel pixel into a grayscale value. All experiments are performed on RTX 3090 GPU. 

The hyper-parameter $T_{TA}$ is set to 100 for backbone GDP, and it is set to 30 for backbone DPS and DDNM. $T_{TC}$ is set to 300 for GDP, 30 for DPS, and DDNM. Higher values of $T_{TA}$ and $T_{TC}$ result in increased computational cost and poorer performance by inaccurate optical flow. $\gamma$ and $\lambda$ are set to 100 and 0.5, respectively. The early stopping sampling strategy is only applied to video super-resolution, denoising, inpainting, and low-light video enhancement on backbone GDP, the hyper-parameter $T_{ES}$ is set to 50. When $T_{ES}$ ranges from 0 to 50, the WE is improved, and the other metrics remain unchanged. After 50, when $T_{ES}$ increases, although WE is improved, the other metrics have significantly deteriorated. 

\section{Ablation Study}

\begin{table}[t]
\centering
\resizebox{0.48\textwidth}{17.5mm}{
\begin{tabular}{ccccccccccc}
\toprule
\multirow{2}{*}{SLTA}        &CNFA            & $\times$   & $\checkmark$ & $\checkmark$ & $\checkmark$    & $\checkmark$ & $\checkmark$  & $\checkmark$     & $\checkmark$    & $\checkmark$  \\\cline{2-11}
                             &SCTA-F          & $\times$   & $\times$     & $\checkmark$ & $\checkmark$    & $\checkmark$ & $\checkmark$  & $\checkmark$     & $\checkmark$    & $\checkmark$  \\ \cline{2-11}  
                             &SCTA-S          & $\times$   & $\times$     & $\times$     & $\checkmark$    & $\checkmark$ & $\checkmark$  & $\checkmark$     & $\checkmark$    & $\checkmark$  \\ \cline{2-11}             
                             &SCTA-H          & $\times$   & $\times$     & $\times$     & $\times$        & $\checkmark$ & $\checkmark$  & $\checkmark$     & $\checkmark$    & $\checkmark$  \\ \hline                              
\multirow{2}{*}{TCG}         &PCG             & $\times$   & $\times$     & $\times$     & $\times$        & $\times$     & $\checkmark$  & $\checkmark$     & $\checkmark$    & $\checkmark$    \\\cline{2-11}
                             &SCG             & $\times$   & $\times$     & $\times$     & $\times$        & $\times$     & $\times$      & $\checkmark$     & $\checkmark$    & $\checkmark$    \\\hline
\multirow{1}{*}{STNS}        &                & $\times$   & $\times$     & $\times$     & $\times$        & $\times$     & $\times$      & $\times$         & $\checkmark$    & $\checkmark$     \\\hline
\multirow{1}{*}{ESSS}        &                & $\times$    & $\times$    & $\times$     & $\times$        & $\times$     & $\times$      & $\times$         & $\times$        & $\checkmark$     \\ \hline
\multicolumn{2}{l}{PSNR$\uparrow$}            & 20.44      & 20.54        & 20.55        & 20.85           & 21.12        & 21.10         & 21.15            & 21.42           & 21.39 \\
\multicolumn{2}{l}{SSIM$\uparrow$}            & 0.5252     & 0.5378       & 0.5381       & 0.5503          & 0.5611       & 0.5603        & 0.5623           & 0.5847          & 0.5843 \\
\multicolumn{2}{l}{FID$\downarrow$ }          & 171.59     & 171.45       & 171.55       & 170.05          & 168.89       & 168.86        & 168.62           & 167.35          & 167.44 \\
\multicolumn{2}{l}{WE($10^{-2}$)$\downarrow$} & 4.0327     & 3.9265       & 3.8346       & 2.9921          & 2.2806       & 1.9655        & 1.9281           & 0.4586          & 0.4234 \\
\multicolumn{2}{l}{FS$\uparrow$}              & 0.8950     & 0.8984       & 0.8964       & 0.9104          & 0.9220       & 0.9289        & 0.9653           & 0.9867          & 0.9885 \\
\multicolumn{2}{l}{OFME$\downarrow$}          & 4.3595     & 4.0403       & 3.9178       & 2.8699          & 1.8654       & 1.4821        & 1.3540           & 0.9972          & 0.9948 \\
\bottomrule
\end{tabular}
}
\caption{Ablation study for SLR Temporal Attention (SLTA), temporal consistency guidance (TCG), spatial-temporal noise sharing (STNS) and early stopping sampling strategy (ESSS) on 4$\times$ video super-resolution task. SLTA module consists Cross-Neighbour-Frame Attention (CNFA) and Self-Corrected Trajectory Attention (SCTA). SCTA-F, SCTA-S, and SCTA-H denote SCTA with only Flow-based, Similarity-based, and Historically-best Trajectory, respectively. TCG module consists of Pixel Consistency Guidance (PCG) and Semantic Consistency Guidance (SCG). }
\label{Ablation}
\end{table}

In this section, we further give a more detailed ablation study. Take video super-resolution as an example, Table \ref{Ablation} lists the quantitative comparison results on evaluation data by adding these modules one by one. In the Self-Corrected Trajectory Attention module, Flow-based Trajectory has little gain due to the inaccurate optical flow. Through self-corrected strategy, Similarity-based and Historically-best Trajectory can bring more gains for all six metrics. For Temporal Consistency Guidance, Pixel-level and Semantic-level Consistency Guidance are more beneficial for WE and FS, respectively.

\section{Comparison with State-of-the-art Methods}

\begin{table}[t]
\centering
%\begin{tabular}{m{5cm}m{0.7cm}<{\centering}m{0.7cm}<{\centering}}
%\begin{tabular}{m{3.0cm}m{2.5cm}<{\centering}m{1.6cm}<{\centering}m{1.3cm}<{\centering}m{1.3cm}<{\centering}m{1.8cm}}
\resizebox{0.49\textwidth}{10mm}{
\addtolength{\tabcolsep}{-5pt}
\begin{tabular}{l|c|c|c|c|c|c}
%\begin{tabular}{m{5cm}m{0.7cm}m{0.7cm} }
%\begin{tabular}
\toprule
Methods                                      & PSNR$\uparrow$  & SSIM$\uparrow$  & FID$\downarrow$ & WE$\downarrow$  & FS$\uparrow$ & OFME$\downarrow$\\
\hline
VRT                                          &\textbf{21.66}      &\textbf{0.6890}       &230.00                &\textbf{2.5082}     &\textbf{0.9782}     &\textbf{0.1953} \\
DPS                                            &20.02               &0.5207                &\underline{220.19}    &6.1475              &0.8356              &11.3684 \\
DPS+ZVRD                                      &\underline{21.59}  &\underline{0.6315}     &\textbf{202.14}        &\underline{2.9470}  &\underline{0.8731}  &\underline{2.0869} \\
\bottomrule
\end{tabular}
}
\caption{Quantitative comparison with state-of-the-art methods for video deblurring (motion blur, blur kernels are 33$\times$33 with a strength of 0.5). The best results are highlighted in bold and the second best results are underlined. WE is expressed as a percentage (\%). VRT is a supervised method, the others are zero-shot methods.}
\label{ComparisonVDB}
\end{table}

\begin{table}[t]
\centering
%\begin{tabular}{m{5cm}m{0.7cm}<{\centering}m{0.7cm}<{\centering}}
%\begin{tabular}{m{3.0cm}m{2.5cm}<{\centering}m{1.6cm}<{\centering}m{1.3cm}<{\centering}m{1.3cm}<{\centering}m{1.8cm}}
\resizebox{0.49\textwidth}{13.0mm}{
\addtolength{\tabcolsep}{-5pt}
\begin{tabular}{l|c|c|c|c|c|c}
%\begin{tabular}{m{5cm}m{0.7cm}m{0.7cm} }
%\begin{tabular}
\toprule
Methods                                       & PSNR$\uparrow$  & SSIM$\uparrow$  & FID$\downarrow$ & WE$\downarrow$  & FS$\uparrow$ & OFME$\downarrow$\\
\hline
FastDVDNet                                       &\textbf{28.40}       &\textbf{0.8443}    &\textbf{112.88}     &\textbf{2.3710}    &\textbf{0.9678}    &\textbf{0.6985}  \\
UDVD                                             &27.29                &0.7986             &192.52              &3.4751             &0.9550             &1.6719   \\                
ZS-N2N                                           &25.33                &0.6749             &267.98              &3.8726             &0.9494             &3.2033   \\
DDNM                                             &28.11	             &0.8285	         &136.73	          &3.2159	          &0.9591	          &0.8738  \\
DDNM+ZVRD                                        &\underline{28.25}    &\underline{0.8296} &\underline{135.22}  &\underline{2.9247} &\underline{0.9604} &\underline{0.7495}  \\
\bottomrule
\end{tabular}
}
\caption{Quantitative comparison with state-of-the-art methods for video denoising (Gaussian noise, $\sigma=50$).  The best results are highlighted in bold and the second best results are underlined. WE is expressed as a percentage (\%). FastDVDNet is a supervised method, UDVD is a unsupervised method, the others are zero-shot methods.}
\label{ComparisonVDN}
\end{table}

\begin{table}[t]
\centering
%\begin{tabular}{m{5cm}m{0.7cm}<{\centering}m{0.7cm}<{\centering}}
%\begin{tabular}{m{3.0cm}m{2.5cm}<{\centering}m{1.6cm}<{\centering}m{1.3cm}<{\centering}m{1.3cm}<{\centering}m{1.8cm}}
\resizebox{0.49\textwidth}{13.0mm}{
\addtolength{\tabcolsep}{-5pt}
\begin{tabular}{l|c|c|c|c|c|c}
%\begin{tabular}{m{5cm}m{0.7cm}m{0.7cm} }
%\begin{tabular}
\toprule
Methods                                      & PSNR$\uparrow$  & SSIM$\uparrow$  & FID$\downarrow$ & WE$\downarrow$  & FS$\uparrow$ & OFME$\downarrow$\\
\hline
RePaint                                      &32.24              &0.9428             &30.47            &3.6251                 &0.9558              &0.7236   \\
DDNM                                         &\underline{32.55}  &\underline{0.9453} &\underline{8.90} &2.4186                 &\underline{0.9821}  &\underline{0.0669}\\
DDNM+ZVRD                                    &\textbf{32.63}     &\textbf{0.9678}    &\textbf{8.36}    &\underline{2.3575}     &\textbf{0.9831}     &\textbf{0.0575} \\
GDP                                          &26.96              &0.8011             &40.99            &2.9366                 &0.9610              &0.3046 \\
GDP+ZVRD                                     &27.13              &0.8067             &35.85            &\textbf{1.6072}        &0.9799              &0.2122 \\
\bottomrule
\end{tabular}
}
\caption{Quantitative comparison with state-of-the-art methods for 25$\%$ video inpainting. The best results are highlighted in bold and the second best results are underlined. WE is expressed as a percentage (\%). All methods are zero-shot methods.}
\label{ComparisonVInp}
\end{table}

Tables \ref{ComparisonVDB}-\ref{ComparisonVInp} list the quantitative results for the video deblurring, video denoising, and video inpainting, respectively. For all tasks, our method can improve the performance in most of the six metrics. For video denoising, our method can boost GDP to outperform the unsupervised method UDVD.

\begin{figure*}
    \centering
    \includegraphics[width=0.9\linewidth]{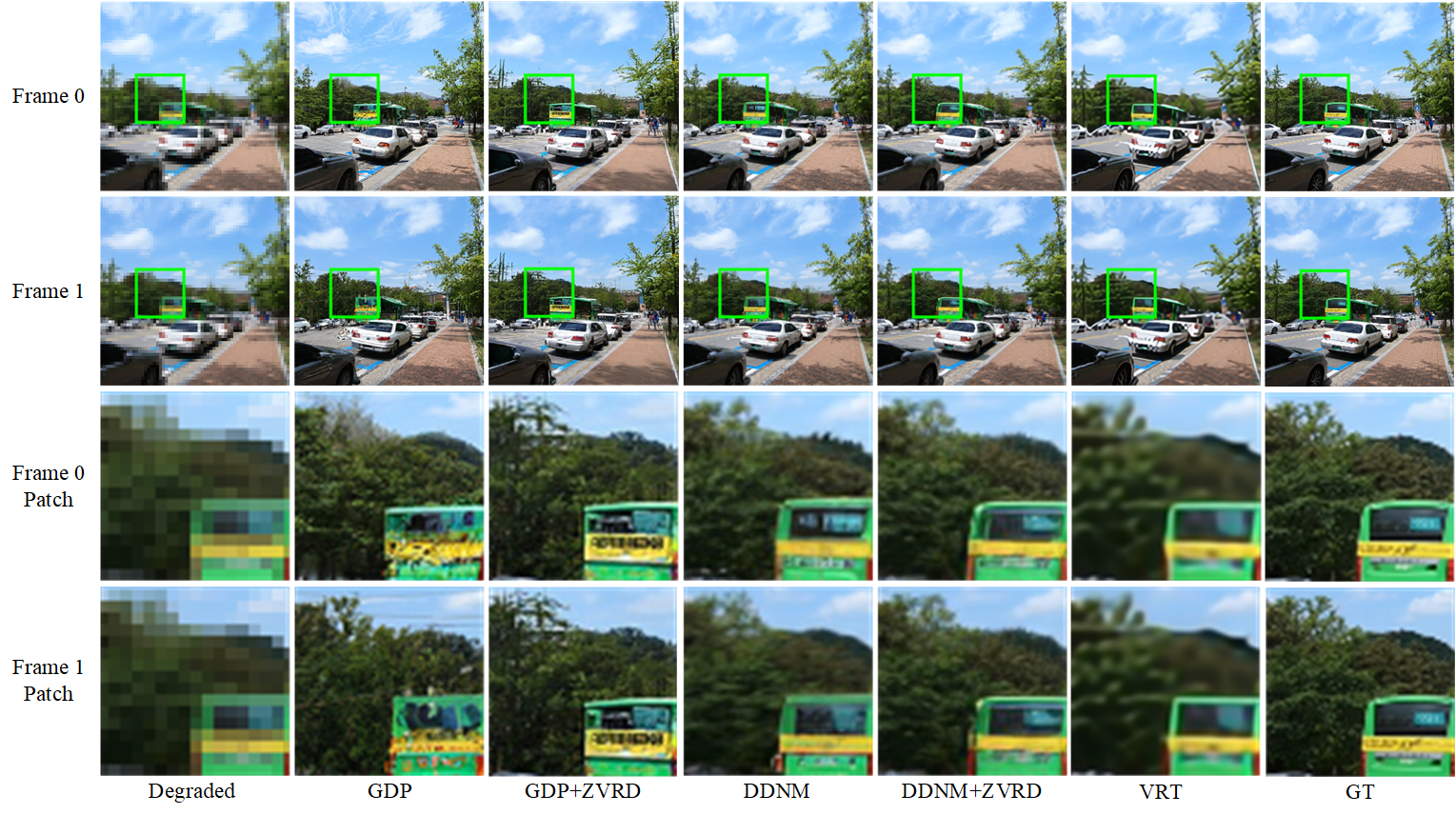}
    \caption{Visual quality comparison for video super-resolution. Zoom in for better observation.}
    \label{fig:videosr}
\end{figure*}

\begin{figure*}
    \centering
    \includegraphics[width=0.9\linewidth]{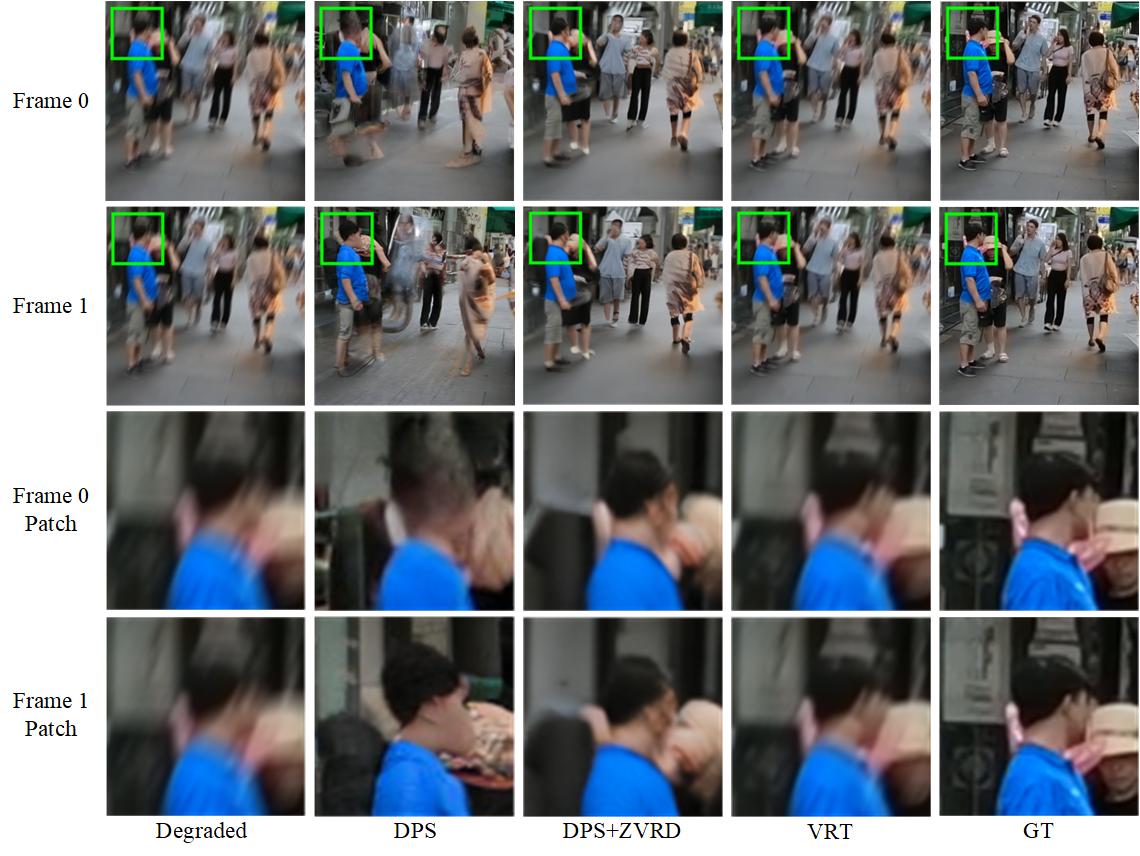}
    \caption{Visual quality comparison for video deblurring. Zoom in for better observation.}
    \label{fig:videodb}
\end{figure*}

\begin{figure*}
    \centering
    \includegraphics[width=0.9\linewidth]{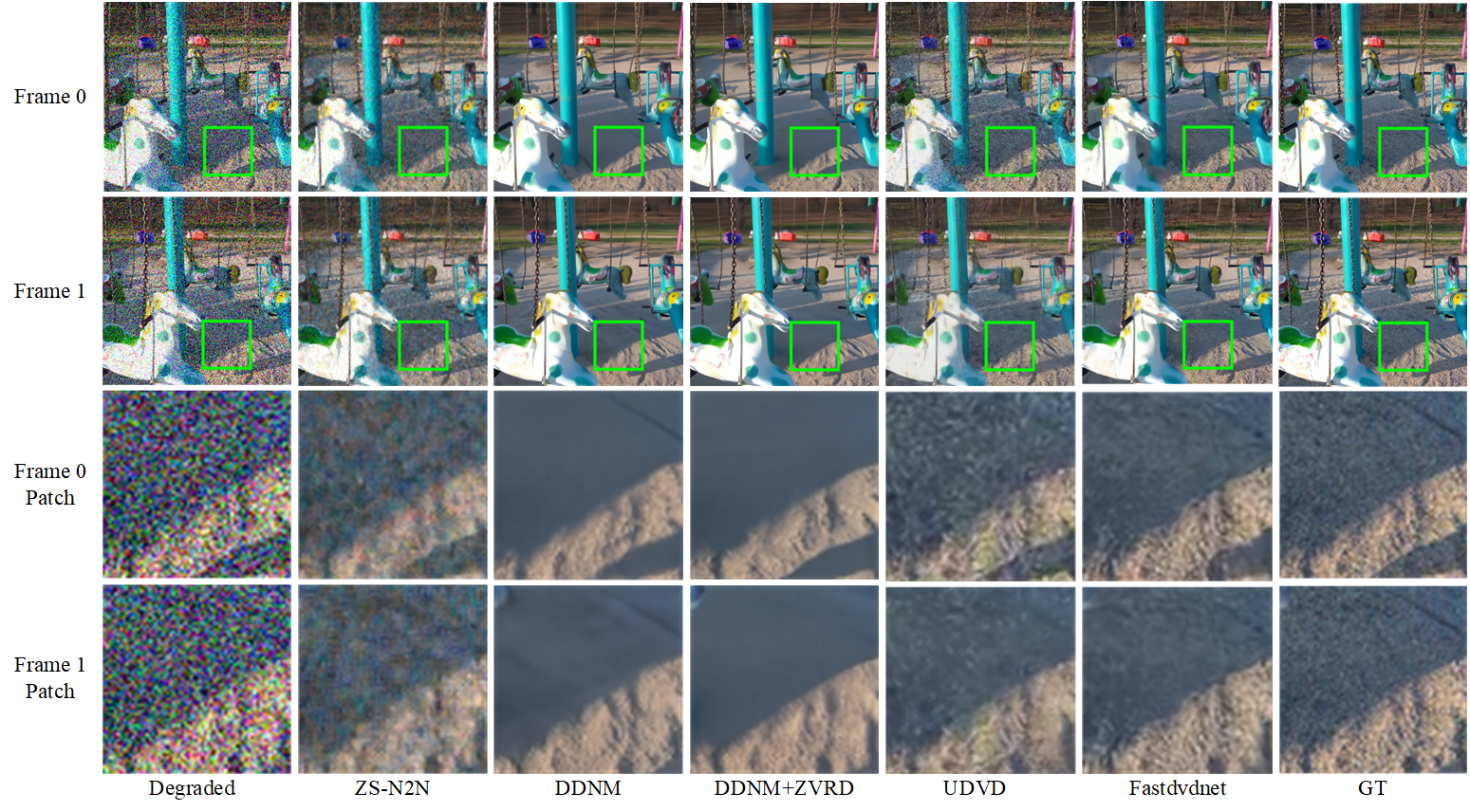}
    \caption{Visual quality comparison for video denoising. Zoom in for better observation.}
    \label{fig:videodn}
\end{figure*}

\begin{figure*}
    \centering
    \includegraphics[width=0.9\linewidth]{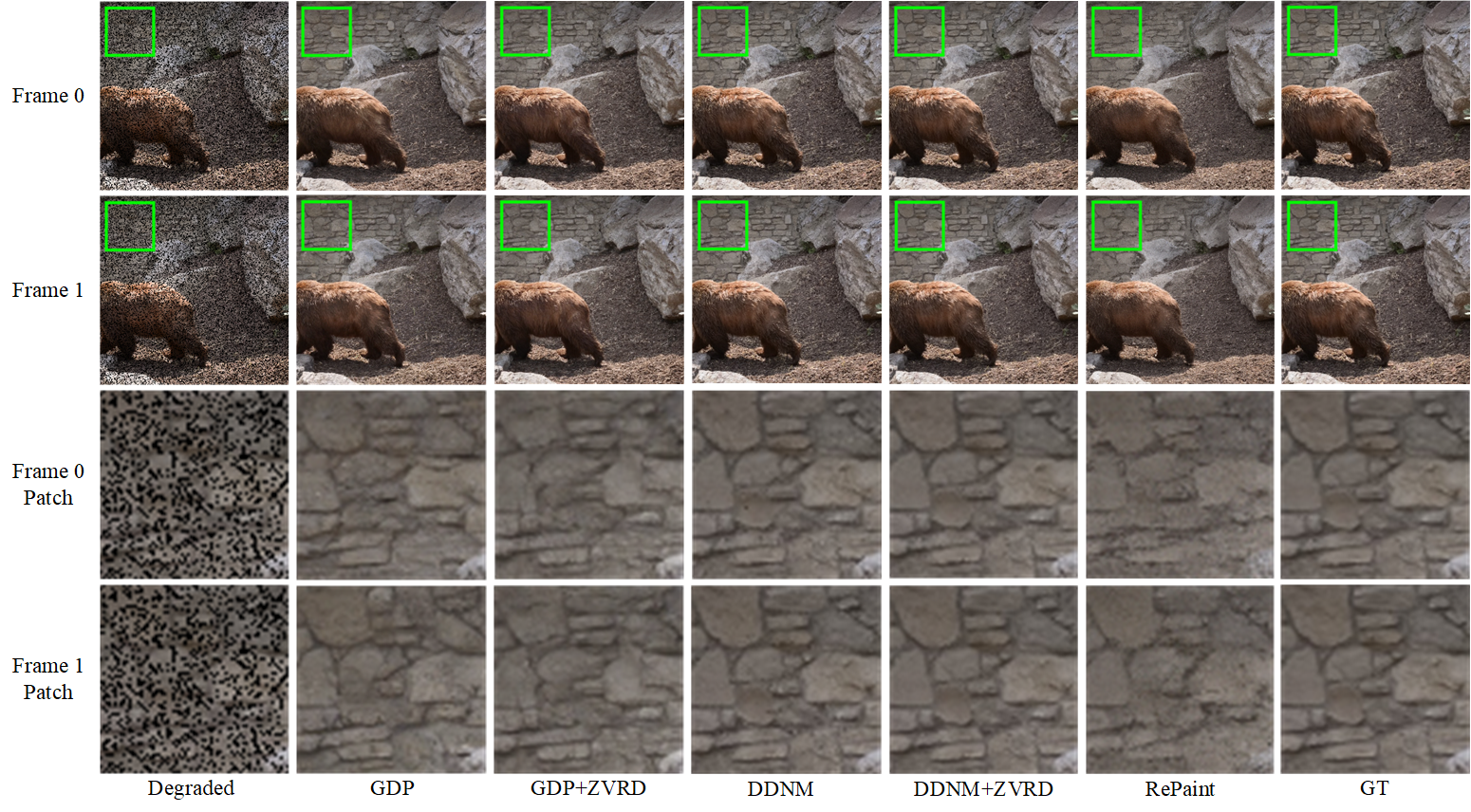}
    \caption{Visual quality comparison for video inpainting. Zoom in for better observation.}
    \label{fig:videoinp}
\end{figure*}

\begin{figure*}
    \centering
    \includegraphics[width=0.9\linewidth]{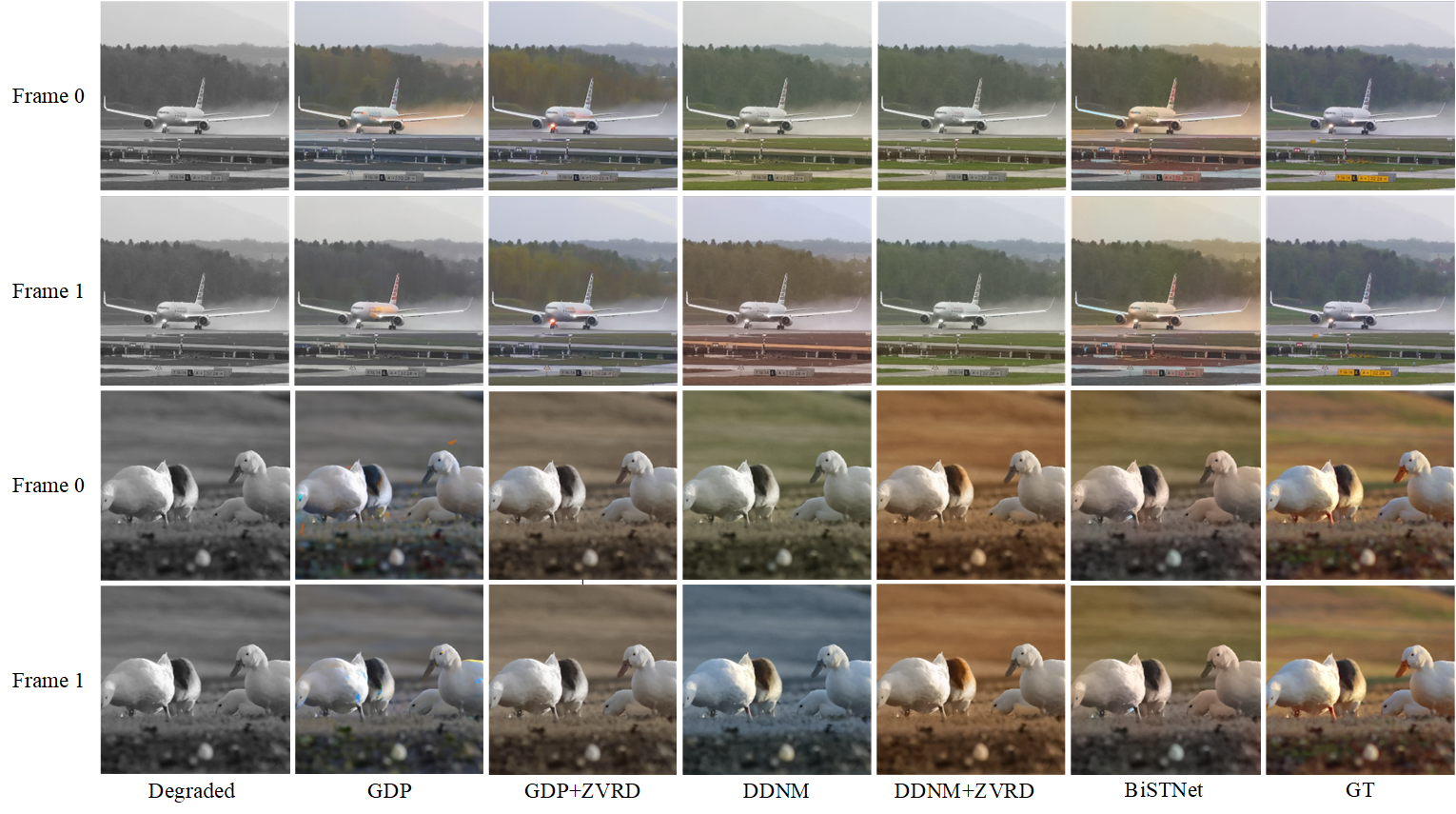}
    \caption{Visual quality comparison for video colorization. Zoom in for better observation.}
    \label{fig:videocolor}
\end{figure*}

\begin{figure*}
    \centering
    \includegraphics[width=0.9\linewidth]{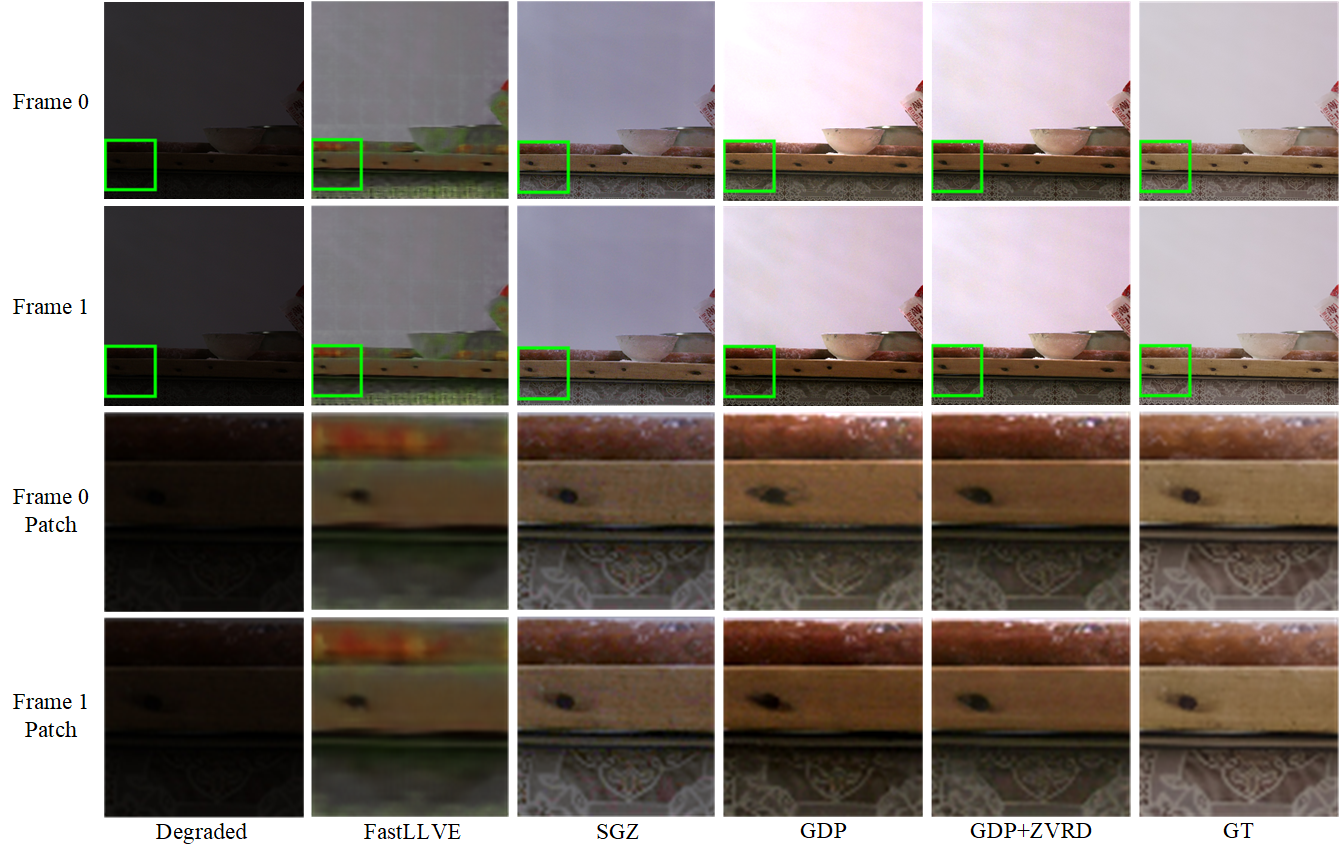}
    \caption{Visual quality comparison for low-light video enhancement. Zoom in for better observation.}
    \label{fig:videolle}
\end{figure*}

\begin{figure*}
    \centering
    \includegraphics[width=0.9\linewidth]{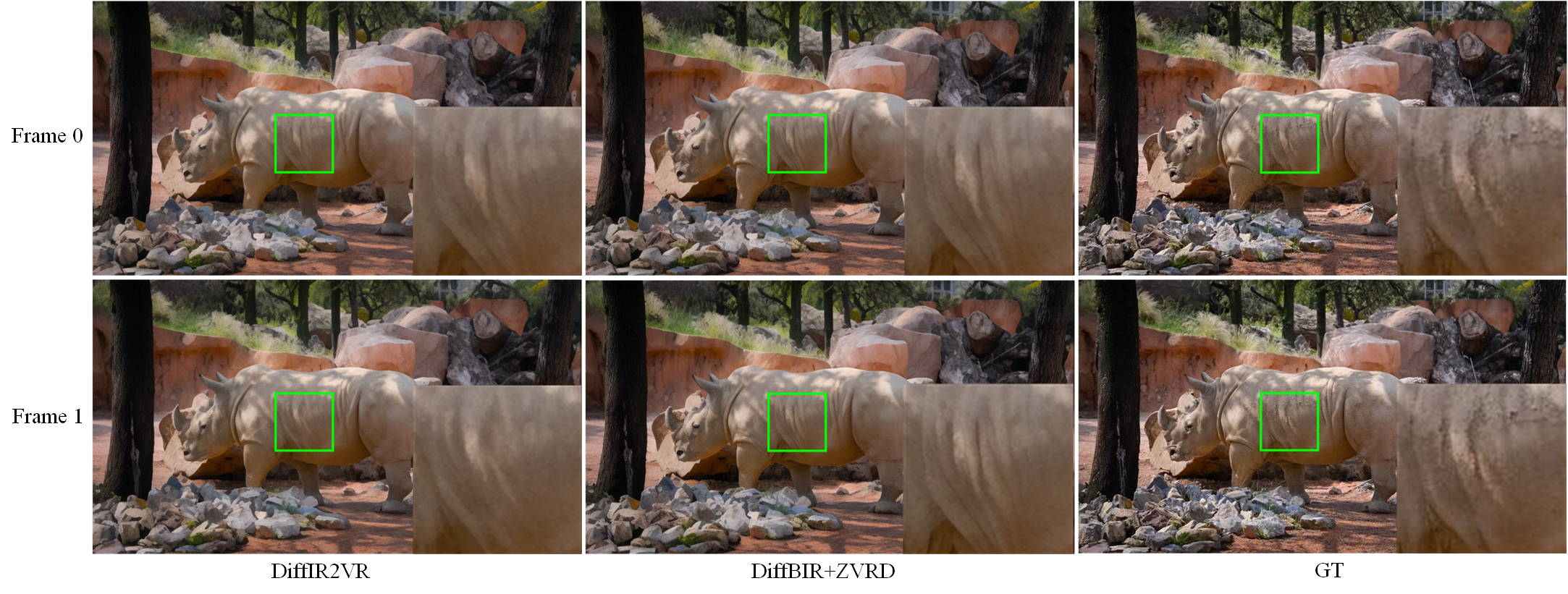}
    \caption{Visual quality comparison for blind video super-resolution. Zoom in for better observation.}
    \label{fig:videobsr}
\end{figure*}

Fig. \ref{fig:videosr}, \ref{fig:videodb}, \ref{fig:videodn}, \ref{fig:videoinp}, \ref{fig:videocolor}, \ref{fig:videolle} present the visual comparison results on the evaluation data for video super-resolution, deblurring, denoising, inpainting, colorization and low-light video enhancement, respectively. It can be observed that our method can improve the temporal consistency on all six tasks. For video super-resolution,
GDP+ZVRD and DDNM+ZVRD restore more details than VRT. For video denoising, ZS-N2N and UDVD still remain noisy, DDNM+ZVRD achieves a better balance of denoising and preserving details. For low-light video enhancement, the results of FastLLVE are blurry and exhibit a color shift (green). The results of SGZ show a bluish color shift and temporal inconsistent color artifacts in the patches. GDP+ZVRD has better results. Fig. \ref{fig:videobsr} presents the visual comparison results on the evaluation data for blind video super-resolution. It can be observed that our method can restore more temporally consistent details than DiffIR2VR. We also present a video demo to further present the temporal consistency of our method.

To further evaluate the temporal consistency, we performed a user study. Fourteen volunteers participate in the subjective test. To ease the comparison, we only compared the results of DDNM and DDNMZVRD on eight groups of videos. The user is asked to evaluate the temporal consistency of the video with a score ranged from 1 to 5, where 5 indicates good quality and 1 indicates bad quality. For fair comparison, the two results for the same degraded video are displayed on the screen simultaneously, with their positions (left or right) assigned randomly. The average score for DDNMZVRD and DDNM is 3.82 and 2.8, respectively, which demonstrates that our method has much better temporal consistency. 

\bibliography{aaai25}